\documentclass[sigconf]{acmart}

\usepackage{appendix}
\usepackage{enumitem}
\usepackage[linesnumbered,ruled,vlined]{algorithm2e}
\usepackage{subfigure}
\usepackage{multirow}
\usepackage{balance}
\usepackage[normalem]{ulem}

\usepackage{amsfonts,amssymb}
\usepackage{amsmath,bm}
\usepackage{threeparttable}

\usepackage{color}
\usepackage{soul}
\usepackage{float}
\usepackage{makecell} 

\usepackage{pifont}
\usepackage{graphicx}
\usepackage{fancyhdr}
\usepackage{longtable}

\SetKwInput{KwInput}{Input}                
\SetKwInput{KwOutput}{Output}              

\newcommand{\smallsection}[1]{\vspace{1mm}\noindent \textbf{#1.}}

\newcommand{\Method}{\textsf{PathFusion}}

\newcommand{\PathFull}{Modality Similarity Path}
\newcommand{\Path}{MSP}
\newcommand{\Fuse}{IRF}
\newcommand{\FuseFull}{Iterative Refinement Fusion}

\newcommand{\DICEWS}[1]{DICEWS{#1}}

\newcommand{\WIKIYAGO}[1]{WY50K{#1}}

\newcommand{\FBDB}{FB15K-DB15K}
\newcommand{\FBYG}{FB15K-YG15K}

\newcommand{\HitNFull}[0]{Hits@$N$}
\newcommand{\HitN}[0]{H@$N$}

\AtBeginDocument{%
  \providecommand\BibTeX{{%
    \normalfont B\kern-0.5em{\scshape i\kern-0.25em b}\kern-0.8em\TeX}}}

\setcopyright{acmcopyright}
\copyrightyear{2023}
\acmYear{2023}
\setcopyright{acmlicensed}
\begin{document}

\title{ Universal Multi-modal Entity Alignment via Iteratively Fusing Modality Similarity Paths}


\author{Bolin Zhu$^1$, Xiaoze Liu$^2$, Xin Mao$^3$, Zhuo Chen$^4$, Lingbing Guo$^4$, Tao Gui$^1$, Qi Zhang$^1$}
\affiliation{
  \institution{$^1$Fudan University}
  \institution{$^2$Purdue University}
  \institution{$^3$Nanyang Technological University}
  \institution{$^4$Zhejiang University}
}
\email{{blzhu20, tgui, qz}fudan.edu.cn, xiaoze@purdue.edu, xin.mao@ntu.edu.sg, {zhuo.chen, lbguo}@zju.edu.cn}

\begin{abstract}

  The objective of Entity Alignment (EA) is to identify equivalent entity pairs from multiple Knowledge Graphs (KGs) and create a more comprehensive and unified KG.
  The majority of EA methods have primarily focused on the structural modality of KGs, lacking exploration of multi-modal information.
  A few multi-modal EA methods have made good attempts in this field. Still,  they have two shortcomings: (1) inconsistent and inefficient modality modeling that designs complex and distinct models for each modality; (2) ineffective modality fusion due to the heterogeneous nature of modalities in EA.
  
  To tackle these challenges, we propose \Method{}, consisting of two main components: (1) \Path{}, a unified modeling approach that simplifies the alignment process by constructing paths connecting entities and modality nodes to represent multiple modalities; (2) \Fuse{}, an iterative fusion method that effectively combines information from different modalities using the path as an information carrier.
  Experimental results on real-world datasets demonstrate the superiority of \Method{} over state-of-the-art methods, with $22.4\%-28.9\%$ absolute improvement on Hits@1, and $0.194-0.245$ absolute improvement on MRR. 
\end{abstract}




\maketitle

\section{Introduction}

Knowledge graphs (KGs) are valuable representations of structured knowledge pertaining to real-world objects, and they play a crucial role in various practical applications, including
 question answering~\cite{YongCaoPlus}, entity~\cite{ner1,typing} and relation extraction~\cite{hu2021semi}. However, existing KGs often suffer from significant incompleteness~\cite{OpenEA2020VLDB} due to their construction from diverse data sources, resulting in overlapping entities across different KGs. This situation presents an opportunity to integrate multiple KGs by considering the overlapped entities.

A commonly employed strategy for KG integration is known as \textbf{Entity Alignment} (EA)\cite{OpenEA2020VLDB}, which aims to align entities from different KGs that refer to the same real-world objects. Given two KGs and a small set of pre-aligned entities (referred to as \emph{seed alignment}), EA identifies all possible alignments between the KGs. Recent studies based on embedding techniques\cite{KECG19, RREA20, AliNet20, HyperKA20, DualAMN21} have demonstrated high effectiveness in performing EA, primarily focusing on the \emph{relational modality} of KGs. These approaches operate under the assumption that the neighbors of two equivalent entities in separate KGs are also equivalent~\cite{AttrGNN20}. By leveraging this assumption, these methods align entities by training an EA model to learn representations of KGs. 

\begin{figure}[t]
    \centering 
    \hspace{-2mm}
    \includegraphics[width=3.3in]{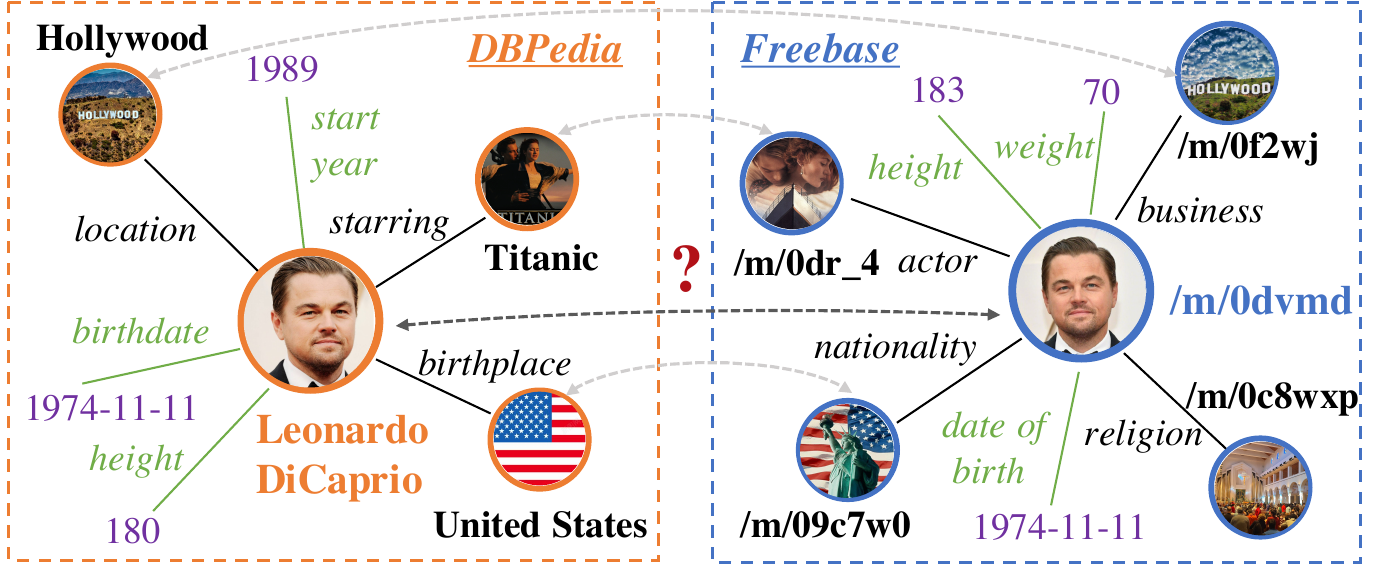} 
    \vspace{-2mm} 
    \caption{An example of multi-modal entity alignment.} 
    \label{fig:intro_demo}
\end{figure}  



Many existing studies on Entity Alignment (EA) face issues with the geometric embedding space, limiting their real-world applicability~\cite{OpenEA2020VLDB, ClusterEA}. To address these problems, incorporating information from "side modalities" of knowledge graphs has been proposed to enhance the robustness and effectiveness of EA~\cite{AttrGNN20, EVA20, AttrE19, TEAGNN21, DualMatch, MCLEA, MSNEA}. Textual information is commonly used, but recent studies have highlighted potential test data leakage issues when incorporating textual information~\cite{AttrGNN20, NoMatch21}. As a result, researchers have explored alternative modalities like visual, attribute, and temporal modalities to improve EA performance~\cite{EVA20, JAPE17, TEAGNN21}.

The field of multi-modal learning has inspired researchers to investigate multi-modal EA, where multiple side modalities are jointly modeled for EA~\cite{MMEA20, ACKMMEA, MCLEA, MSNEA}. However, these methods have certain limitations:

\begin{itemize}[topsep=0pt,itemsep=0pt,parsep=0pt,partopsep=0pt,leftmargin=*]
    \item \textbf{Inconsistent and inefficient modality modeling.}
      Different models are used for different modalities, leading to time-consuming and non-generalizable approaches. Additionally, these methods mainly focus on learning vectorized representations and require transformation into similarity scores between entities.
        \item  \textbf{Ineffective modality fusion.} 
      Complex learning procedures are employed to integrate information from different modalities. However, the heterogeneous nature of modalities in knowledge graphs poses challenges to effective integration. An example can illustrate this point.
    \end{itemize}

\begin{example}
  In classic multi-modal (MM) learning scenarios, different modalities are typically homogeneous. However, in the context of Multi-Modal Entity Alignment (MMEA), different modalities are heterogeneous. As shown in Fig.~\ref{fig:compare_demo}, the texts provide explicit descriptions of the images. However, in knowledge graphs (KGs), the relations, attributes, and images exhibit heterogeneity. For instance, in the MMKG example depicting Michael Jordan, we cannot infer his height, weight, or birth date from the image alone, nor can we infer the image from the attributes. Consequently, modeling interactions between different modalities becomes challenging and unnecessary.
\end{example}

To address these challenges, we propose a novel method called \Method{} for multi-modal EA. We introduce the \emph{\PathFull{}} (\Path{}) approach to model different modalities in a unified manner. Instead of embedding modality information, \Path{} constructs a \emph{path} between entities by connecting bipartite graphs, simplifying the alignment process and enabling direct transformation of information into alignment results. We also propose the \emph{\FuseFull{}} (\Fuse{}) method to effectively fuse information from different modalities. Unlike existing methods, \Fuse{} iteratively fuses information into a unified \emph{path} to make it more discriminative, rather than forcing interactions between modalities. 
To summarize, our contributions are as follows:

\begin{figure}[t]
    \centering 
    \hspace{-2.8mm}
    \includegraphics[width=3.4in]{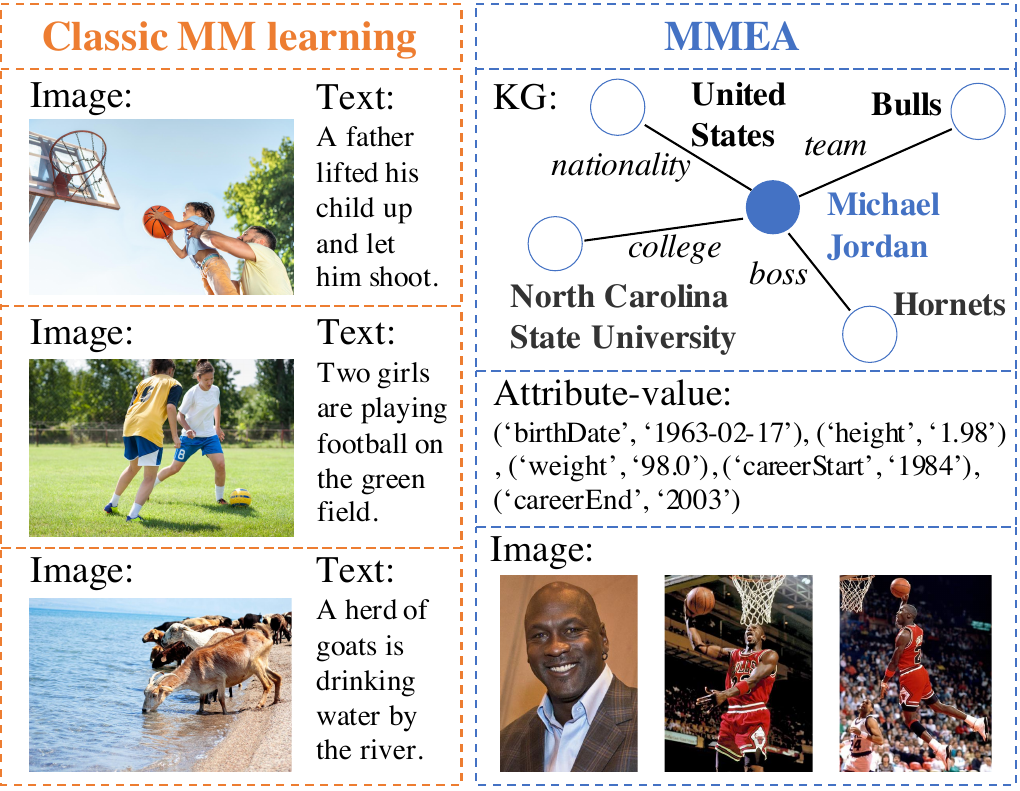} 
    \vspace{-4mm}
    \caption{Comparison between multi-modal entity alignment and the classic multi-modal learning tasks.}
    \label{fig:compare_demo}
    \vspace{-6mm}
\end{figure}

\begin{itemize}[topsep=0pt,itemsep=0pt,parsep=0pt,partopsep=0pt,leftmargin=*]
\item {\textbf{Model}. 
  We propose the \Method{}\footnote{\href{https://github.com/blzhu0823/PathFusion}{https://github.com/blzhu0823/PathFusion}} framework for multi-modal EA, which models different modalities in a unified way and effectively fuses their information.
}
\item {\textbf{Path.} 
We introduce the \Path{} approach to model different modalities in a unified manner, allowing for direct alignment results without information loss.
}
\item {\textbf{Fuse.} 
We propose the \Fuse{} method to iteratively fuse information from different modalities using the \emph{path} as the information carrier.
}
\item {\textbf{Experiments.} 
We conduct extensive experiments on real-world datasets, demonstrating that \Method{} outperforms state-of-the-art methods by a large margin.
}

\end{itemize}


\section{Related Work}
\label{sec:related_work}

In this section, we review the related work of EA, including structure-based EA and multi-modal EA.

\subsection{Structure-based Entity Alignment}

The foundation of EA is to learn the graph relation structures of KGs. Existing structure-based EA models can be divided into two categories: \emph{KGE}-\emph{based methods}\cite{MTransE17, IPTransE17, BootEA18, AttrE19} and \emph{GNN-based methods}\cite{GCN-Align18, KECG19, MRAEA20, AliNet20, HyperKA20, DualAMN21}.
The former uses the KG embedding models (e.g., TransE~\cite{TransE13}) to learn entity embeddings, while the latter uses GNNs~\cite{GCN17,yin2022dgi}.
\Method{} is designed for multi-modal EA, and thus can incorporate any structure-based EA model as the backbone.

\subsection{Multi-Modal Entity Alignment}

All existing proposals of multi-modal EA incorporate \emph{side modalities} of KGs, including \emph{textual modality} of entity names and descriptions \cite{EASY21, SEU21, li2023vision}, \emph{attribute modality} that use attributes of entities~\cite{AttrE19,AttrGNN20, ACKMMEA}, \emph{temporal modality} that uses timestamps of relations~\cite{TEAGNN21, TREA22,DualMatch}, and \emph{visual modality} \cite{EVA20, MSNEA, MCLEA} that use images of entities. Some of such proposals are able to perform EA even without seed alignment \cite{EASY21, SEU21}.
However, models that incorporate textual modality may be overestimated due to potential test data leakage~\cite{JEANS20, AttrGNN20, EVA20, NoMatch21}. 
Thus, other modalities are more reliable for EA. Recently, some multi-modal EA models have been proposed to incorporate multiple modalities~\cite{MMEA20, ACKMMEA, MSNEA, MCLEA}. However, they face limitations in incorporating all modalities due to the lack of a unified framework. Also, their performance is restricted by the absence of an effective mechanism to fuse multiple modalities. In contrast, our proposed method, \Method{}, stands out as the first approach to successfully incorporate all modalities and achieve state-of-the-art performance. 

\begin{table}\small
    \centering
    \setlength{\tabcolsep}{2mm}{
    \begin{tabular}{lcccc}
    \toprule
    \textbf{Method} & \textbf{Relation} & \textbf{Visual} & \textbf{Attribute} & \textbf{Temporal}\\
    \hline
    MTransE~\cite{MTransE17} & \ding{51} &  \ding{55} & \ding{55} & \ding{55} \\
    IPTransE~\cite{IPTransE17} & \ding{51} &  \ding{55} & \ding{55} & \ding{55} \\
    TransEdge~\cite{TransEdge19} & \ding{51} &  \ding{55} & \ding{55} & \ding{55} \\
    MRAEA~\cite{MRAEA20} & \ding{51} &  \ding{55} & \ding{55} & \ding{55} \\
    RREA~\cite{RREA20} & \ding{51} &  \ding{55} & \ding{55} & \ding{55} \\
    DualAMN~\cite{DualAMN21} & \ding{51} &  \ding{55} & \ding{55} & \ding{55} \\
    JAPE~\cite{JAPE17} & \ding{51} &  \ding{55} & \ding{51} & \ding{55} \\
    GCN-Align~\cite{GCN-Align18} & \ding{51} &  \ding{55} & \ding{51} & \ding{55} \\
    AttrE~\cite{AttrE19} & \ding{51} &  \ding{55} & \ding{51} & \ding{55} \\
    AttrGNN~\cite{AttrGNN20} & \ding{51} &  \ding{55} & \ding{51} & \ding{55} \\
    TEA-GNN~\cite{TEAGNN21} & \ding{51} &  \ding{55} & \ding{55} &  \ding{51}  \\
    TREA~\cite{TREA22} & \ding{51} &  \ding{55} & \ding{55} &  \ding{51}  \\
    STEA~\cite{cai2022simple} & \ding{51} &  \ding{55} & \ding{55} &  \ding{51}  \\
    DualMatch~\cite{DualMatch} & \ding{51} &  \ding{55} & \ding{55} &  \ding{51}  \\
    MMEA~\cite{MMEA20} & \ding{51} &   \ding{51}  &  \ding{51} &    \ding{55}   \\
    EVA~\cite{EVA20} & \ding{51} &   \ding{51}  &  \ding{51} &    \ding{55}   \\
    MSNEA~\cite{MSNEA} & \ding{51} &   \ding{51}  &  \ding{51} &    \ding{55}   \\
    MCLEA~\cite{MCLEA} & \ding{51} &   \ding{51}  &  \ding{51} &    \ding{55}   \\
    ACK-MMEA~\cite{ACKMMEA} & \ding{51} &   \ding{51}  &  \ding{51} &    \ding{55}   \\
    MEAformer~\cite{MEAformer} & \ding{51} &   \ding{51}  &  \ding{51} &    \ding{55}   \\
    \textbf{Ours} & \ding{51} &   \ding{51}  &  \ding{51} &   \ding{51}   \\
    \bottomrule
    \end{tabular}}
    \caption{A table of methods and their supported modalities. The four modalities are relation, visual, attribute, and temporal.} 
    \label{tab:modalities}
\end{table}

We report the supported modalities of \Method{} with the compared methods in Table~\ref{tab:modalities}. Our method is the only one that can support all modalities.

\section{Problem definition}

In this section, we present our problem definition.

A \textbf{multi-modal knowledge graph} (MMKG) can be denoted as $G = (E,R,Q,T,I,P,K,V,A)$, where $E$ is the set of entities, $R$ is the set of relations.
$T,I,P,K, V$ are sets of items in the side modalities including the temporal, attribute and visual modalities, and can be optional if the dataset do not hold information of such modality. 
$T$ is a set of timestamps that represent the information of the temporal modality. $I$ is the set of images and $P= \{e, i\}~|~e \in E, i \in I$ is the set of entity-image pairs, representing that $i$ is an image of the entity $e$. $K, V, A$ are the set of attribute names, values and entity-attribute triples. $A$ is represented as $\{(e, k, v )~|~e \in E, k \in K, v \in V \}$. 
$Q=\{(h,r,t, \tau)~|~h,t \in E, r \in R, \tau \in T \}$ is the set of quadruples, each of which represents that the subject entity $h$ has the relation $r$ with the object entity $t$  during the time interval $\tau$. $\tau$ here is optional, if not given, the tuple $(h,r,t)$ is often called a triple. For simplicity, we use the term ``triple'' with or without the time interval $\tau$ to represent the quadruple $(h,r,t, \tau)$.

\smallsection{Multi-modal entity alignment} (MMEA) is the process of finding a 1-to-1 mapping of entities $\phi$ from a source MMKG\,$G_s = (E_s,R_s,Q_s,T_s,I_s, P_s,K_s, V_s, A_s)$ to a target MMKG $G_t = (E_t,R_t,Q_t,T_t,$ $I_t, P_t,K_t, V_t, A_t)$.
$\phi = \{(e_s, e_t) \in E_s \times E_t~|~e_s \equiv e_t\}$ , where $e_s \in E_s$, $e_t \in E_t$, and $\equiv$ is an equivalence relation between two entities.
Here, since the timestamps (if available) are overlapped between knowledge graphs, we have $T_s = T_t$. 
A small set of equivalent entities $\phi^{\prime} \subset \phi$ is known beforehand, and is used as training seed alignment.

\section{Methodology}
\label{sec:method}

In this section, we present the proposed \Method{} for EA on multi-modal KGs. First, we give an overview of the proposed \Method{}

\subsection{Overview}

\begin{figure*}[t]
    \centering
    
    \includegraphics[width=6.8in]{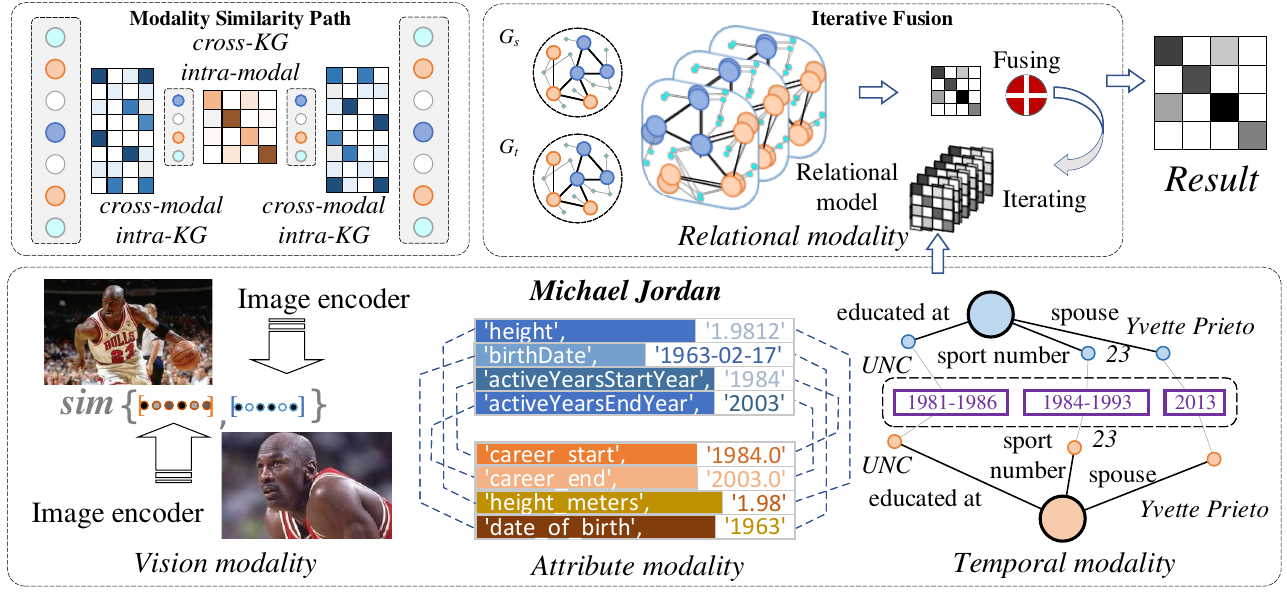}
    \vspace*{-3mm}

    \caption{The proposed \Method{} framework.}
    \label{fig:framework}
\end{figure*}

Fig.~\ref{fig:framework} shows the overview of the proposed \Method{}.
\Method{} consists of two parts: \emph{Universal Modality Modeling} with \PathFull{} and \emph{\FuseFull{}}. \Method{} first aggregates the information of modalities into entity representations by \Path{}, and then iteratively fuse the entity representations from the source and target KGs to obtain the mapping matrix with \Fuse{}. We detail the two parts in the following subsections.

\subsection{Universal Modality Modeling with \PathFull{}}

Previous research has focused on learning vectorized representations from side modalities to enhance the performance of entity alignment (EA). Side modalities provide additional information, as latent relations exist between items of side modalities in two knowledge graphs. For example, in the visual modality, latent relations involve visual similarities between entity images (e.g., two images of the same person are more similar than two images of different people). In the attribute modality, latent relations consist of attribute name and value similarities (e.g., two people with the same height and birth date are more likely to be the same person than two people with different heights and birth dates).

Multi-modal learners incorporate modalities into entity embeddings, which serve as intermediate representations in EA models. Typically, embedding learners aggregate side modality information into entity representations, leveraging explicitly defined entity-side modality relations. For instance, in the visual modality, entity-image pairs $P$ indicate image-entity associations, allowing the concatenation of visual features of entity images to obtain entity embeddings~\cite{EVA20}.

The final output of an EA model is a mapping matrix $M \in \mathbb{R}^{E_s \times E_t}$ representing entity pairs between two knowledge graphs. This matrix is obtained by taking the dot product of entity embeddings from the source and target KGs, formulated as $M = \mathbf{H}_s \mathbf{H}_t^T$, where $\cdot^T$ denotes the transpose operation. Here, $\mathbf{H}_s \in \mathbb{R}^{E_s \times d}$ and $\mathbf{H}_t \in \mathbb{R}^{E_t \times d}$ are the entity embeddings of the source and target KGs, respectively.

Instead of learning entity embeddings, we propose a more direct and efficient approach called \PathFull{} (\Path{}) to model side modalities. \Path{} represents the relations between entities in the source and target KGs by using side modalities as bridges. Each side modality in EA has two sets of items representing the source and target graphs. For example, in the temporal modality, there are two sets of timestamps $T_s$ and $T_t$ representing temporal information in the source and target KGs, respectively. Hence, we first define the similarity between these two sets as a cross-KG bridge.

Assuming we have an arbitrary modality $X$, \Path{} collects the \emph{cross-KG intra-modality similarity} $M_{X}^x \in \mathbb{R}^{|X_s \times X_t|}$ between items in the source and target KGs for the $X$ modality. Then, using the \emph{intra-KG cross-modality similarity} matrices $M_{X}^s \in \mathbb{R}^{|E_s \times X_s|}$ and $M_{X}^t \in \mathbb{R}^{|E_t \times X_t|}$, \Path{} defines the relations between entities and modality items in the source and target KGs. Finally, \Path{} derives the mapping matrix $M_X \in \mathbb{R}^{|E_s \times E_t|}$ by applying a matrix-multiplication-like operator on the similarity matrices. For each $m_{i,j} \in M_X$, we can express \Path{} as:

\begin{equation}
    m_{i,j} =   \underset{x_s \in X_s, x_t \in X_t}{\Psi }\left( M_{X_{i, x_s}}^s M_{X_{x_s, x_t}}^x  M_{X_{j, x_t}}^t\right)
\end{equation}
 
\noindent
The aggregation function $\Psi$ in \Path{} combines similarity matrices, which can be either the max or sum function. If $\Psi$ is the sum function, \Path{} is equivalent to the matrix multiplication operator, defined as $M_X = M_{X}^s M_{X}^x (M_{X}^t)^T$. If $\Psi$ is the max function, \Path{} behaves like a max-pooling operator.

It is evident that \Path{} is a versatile operator applicable to any modality. We only need to specify the cross-KG intra-modality similarity and intra-KG cross-modality similarity for each modality. Subsequent sections provide a comprehensive description of \Path{} for each specific modality.

\smallsection{Relational modality}
In the relational modality, our \Method{} framework incorporates existing relational learning methods~\cite{DualAMN21}. 
\Method{} adopts the state-of-the-art EA method~\cite{DualAMN21} to build a $L$-layer GNN model for learning the representation of the knowledge graph structure.
We train a relational learning model on the two knowledge graphs using the seed alignment as the training set. This yields vectorized entity representation $\mathbf{H} \in \mathbb{R}^{|E| \times d}$, where $d$ is the embedding space dimension. As the modality nodes are entities themselves, the cross-KG intra-modality similarity is the identity matrix $I_d$. The intra-KG cross-modality similarity consists of entity embeddings of the source and target KGs. The mapping matrix $M_R$ is obtained by taking the dot product of entity embeddings from both KGs: $M_R = \mathbf{H}_s I_d (\mathbf{H}_t)^{T}$.

\smallsection{Visual Modality}
In MMKG, entities have visual representations as images. To measure the visual cross-KG similarity, we utilize the visual features extracted from the images using SwinTransformer~\cite{SwinTransformer}. Let $\mathbf{I} \in \mathbb{R}^{|I|\times d_v}$ represent the visual features, where $d_v$ is the dimension of the visual feature space. The visual cross-KG similarity is computed as $M_{I}^x = \mathbf{I}_s (\mathbf{I}_t)^{T}$. The visual intra-KG similarity is derived from the entity-image pairs $P$, where $m_{e_i, j} \in M_I$ between an entity $e_i \in E$ and an image $ j \in I$ would be

\begin{equation}
    m_{e_i,j}= \begin{cases}
    1, & \text{if } (e_i, j) \in P\\
    0, & \text{otherwise}
    \end{cases}
 \end{equation}
 Finally, \Path{} uses the max operator to obtain the mapping matrix $M_I$.


\smallsection{Attribute Modality}
In the attribute modality, the cross-KG intra-modality similarity is obtained by combining attribute name and attribute value similarities. We define the attribute item set $S$ as the set of distinct attribute name-value pairs. The attribute name similarity $M_K$ is computed using a pre-trained language model, where we use the representation of the \texttt{[CLS]} token as the attribute name representation. The attribute value similarity $M_V$ is calculated as the reciprocal of the absolute difference between attribute values. For each $m_{i,j} \in M_V$, we have $m_{i,j} = \frac{1}{|v_i - v_j|}$, where $v_i\in V_s$ and $v_j\in V_t$ are the attribute values of the $i$-th and $j$-th items, respectively. The attribute similarity $M_A^x$ is obtained by multiplying the attribute name similarity and value similarity. The attribute intra-KG similarity is derived from the entity-attribute triples $A$ using the same approach as in the visual modality. The mapping matrix $M_A$ is computed by summing the fused similarities using the sum operator.

\smallsection{Temporal modality} 
Temporal modality is represented as timestamps defined on the relations between entities. Those timestamps are universal across all KGs, which means that for any two MMKGs with temporal information $G_s$ and $G_t$, we have $T_s = T_t$. Therefore, the cross-KG intra-modality similarity is the identity matrix $I_{|T|}$. We can follow~\cite{DualMatch} to obtain the intra-KG similarity.  
Specifically, we rely on two adjacency matrices extracted from the MMKGs: (i) entity-entity matrix $\boldsymbol{A}$, built following~\cite{LightEA}; and (ii) entity-timestamp matrix $\boldsymbol{A^t}$, built by counting the number of relations that one entity $e$ and one timestamp $\tau$ appear together. We follow~\cite{DATTI22, LightEA, DualMatch} to conduct an $L$-layer graph convolution-like forward pass to generate an aggregated feature matrix: $\boldsymbol{\Theta} = \left[\boldsymbol{A} \boldsymbol{A^t} || \boldsymbol{A}^{2} \boldsymbol{A^t} || \ldots || \boldsymbol{A}^{L} \boldsymbol{A^t}\right]$. Here, $L$ serves as a hyper-parameter that specifies the number of layers for the graph convolution-like forward pass, and $\boldsymbol{A}^{l} = \prod_{i=1}^{l} \boldsymbol{A}$ represents the $l$-hop adjacency matrix of the graph $G$. Finally, we use the sum operator to obtain the mapping matrix $M_T = \boldsymbol{\Theta}_s I_{|T|} \boldsymbol{\Theta}_t^T.$
Note that in many Knowledge Graphs, attributes may include temporal information. For instance, certain attribute pairs can take the form of (``United States," ``was founded," ``July 4, 1776"). However, it's important to emphasize that this temporal information is primarily handled within the attribute modality encoders. Reanalyzing it would be duplicative. Consequently, our focus on temporal information is exclusively centered on the relational facts within the KG.

Overall, \Path{} provides a more direct and efficient way to model the side modalities, allowing for effective entity alignment in MMKG.

\subsection{Iterative Fusing of Multiple Modalities}

Directly fusing the information of side modalities to produce the final output is challenging. Some literature~\cite{MSNEA} suggests letting different modalities interact with each other to strengthen "weak" modalities. However, due to the heterogeneous nature of side modalities in EA, learning the interactions between them is inefficient. Instead, the goal should be to make entities more discriminative by leveraging multiple modalities.

\smallsection{Fusing the Modality Paths}
A practical approach is to fuse the information of multiple modalities by summing them up~\cite{CEAFF20}. To enhance discriminability, we adopt the optimal transport (OT) plan~\cite{Sinkhorn13} instead of a similarity matrix to represent the final alignment matrix. In \Path{}, we have obtained mapping matrices for each modality. We directly sum them to obtain the final mapping matrix $M$ using the following equation:

\begin{equation}
\label{eq:obtain_m}
M = \operatorname{Sinkhorn}(\sum_{X \in \mathcal{X}} M_X)
\end{equation}

\noindent
where $\mathcal{X}$ represents the collection of all modalities. We use the Sinkhorn algorithm~\cite{Sinkhorn13} to obtain the optimal transport plan by iterating $k$ steps to scale the similarity/cost matrix. 
The Sinkhorn algorithm~\cite{Sinkhorn13} is defined as follows:
\begin{equation}
\begin{aligned}
&\qquad S^{0}(\boldsymbol{X})=\exp (\boldsymbol{X}) \\
&\qquad S^{k}(\boldsymbol{X})=\mathcal{N}_{c}\left(\mathcal{N}_{r}\left(S^{k-1}(\boldsymbol{X})\right)\right) \\
&\operatorname{Sinkhorn}(\boldsymbol{X})=\lim _{k \rightarrow \infty} S^{k}(\boldsymbol{X}),
\end{aligned}
\label{eq:sinkhorn_it}
\end{equation}
where $\mathcal{N}_{r}(\boldsymbol{X})=\boldsymbol{X} \oslash\left(\boldsymbol{X} \mathbf{1}_{N} \mathbf{1}_{N}^{T}\right)$ and $\mathcal{N}_{c}=\boldsymbol{X} \oslash$ $\left(\mathbf{1}_{N} \mathbf{1}_{N}^{T} \boldsymbol{X}\right)$ are the row and column-wise normalization operators of a matrix, $\oslash$ represents the element-wise division, and $\mathbf{1}_{N}$ is a column vector of ones. 
In reality, we set the iteration step $k$ to a constant instead of $+\infty$.

\smallsection{Iterative Refinement of Relational Modality}
In the relational modality, we trained a model to obtain the similarity matrix $M_R$. However, this process only uses a small fraction of the total alignment pairs as training seeds (usually around $10\%-30\%$), which may underutilize the information of the relational modality. To address this, we propose to refine the relational modality using the mapping matrix $M$ obtained from the previous step, following the bi-directional strategy~\cite{MRAEA20, RREA20}.
In each iteration, we derive the bi-directional alignment matrices $M$ and $M^\prime$ by applying Eq.~\ref{eq:obtain_m} in both directions. Then, we extract pseudo-seeds from these matrices. Specifically, we consider $(e_i, e_j)$ as a valid seed pair if $e_i \in E_s$, $e_j \in E_t$, $\arg \max_{j^{\prime}}{M_{ij^{\prime}}} = j$, and $\arg \max_{i^{\prime}}{M^{\prime}_{ji^{\prime}}} = i$.
Finally, we fine-tune the relational model using the generated pseudo-seed alignment. This refinement process also enables unsupervised relational learning by using the mapping matrix (fused without $M_R$) as the pseudo labels.

\smallsection{Discussions}
This refinement process differs from the aforementioned "interactions" between modalities because "weak" modalities lack sufficient discriminative information. On the other hand, knowledge graphs possess strong graph structure information and are not considered "weak." The refinement process aims to complete the seed alignment and allows the relational modality to better express its information.

\section{Experiments}

\begin{table}[t]
    \centering
    \caption{Statistics of the multi-modal EA datasets.}
    \vspace{-3mm}
    \label{tab:datasets}
    \setlength{\tabcolsep}{1mm}{
\begin{tabular}{cccccc}
    \Xhline{1px} Dataset & Entity & $\begin{array}{c}\text { Relation } \\
    \text { Triple }\end{array}$ & $\begin{array}{c}\text { Attribute } \\
    \text { Triple }\end{array}$ & Image & Seed \\
    \Xhline{1px} FB15K & 14,951 & 592,213 & 29,395 & 13,444 & - \\
    DB15K & 12,842 & 89,197 & 48,080 & 12,837 & 12,846 \\
    YG15K & 15,404 & 122,886 & 23,532 & 11,194 & 11,199 \\
    \Xhline{1px}
    \end{tabular}
    } 
\end{table}

\begin{table}[t]
    \centering
    \caption{Statistics of \DICEWS{} and \WIKIYAGO{}.}
    \label{tab:time_dataset}
    \vspace{-4mm}
    \setlength{\tabcolsep}{1.4mm}{
    \begin{tabular}{l|c|c|c|c}
    \Xhline{1px}
    Dataset & Entity  & Rels & Time & Triples  \\ \Xhline{1px}
    \DICEWS{}       & 9,517-9,537   & 247-246     & 4,017   & 307,552-307,553 \\
    \WIKIYAGO{} & 49,629-49,222 & 11-30       & 245    & 221,050-317,814 \\
    \Xhline{1px}
    \end{tabular}
    }
    \vspace{-3mm}
    \end{table}

We report on extensive experiments aiming at evaluating the performance of \Method{}.

\begin{table*}[t]\small
    \caption{The  results of \Method{} and the baselines on the \FBDB{} and \FBYG{} datasets.}
    \label{tab:main_table}
    \vspace{-3mm}
\centering
    \setlength{\tabcolsep}{2.7mm}{
\begin{tabular}{l|rrrrr|rrrrr}
    \Xhline{1px} \multicolumn{1}{c|}{\multirow{2}{*}{{Method}}} & \multicolumn{5}{c|}{{FB15K-DB15K}} & \multicolumn{5}{c}{{FB15K-YG15K}} \\ 
     & {H@1}$\uparrow$  & {H@5}$\uparrow$  & {H@10}$\uparrow$  & {MR}$\downarrow$  & {MRR}$\uparrow$  & {H@1}$\uparrow$  & {H@5}$\uparrow$  & {H@10}$\uparrow$  & {MR}$\downarrow$  & {MRR}$\uparrow$  \\ \Xhline{1px}
    MTransE~\cite{MTransE17} & 0.4 & 1.4 & 2.5 & 1239.5 & 0.014 & 0.3 & 1.0 & 1.8 & 1183.3 & 0.011 \\
    IPTransE~\cite{IPTransE17} & 4.0 & 11.2 & 17.3 & 387.5 & 0.086 & 3.1 & 9.5 & 14.4 & 522.2 & 0.070 \\
    GCN-Align*~\cite{GCN-Align18} & 7.1 & 16.5 & 22.4 & 304.5 & 0.106 & 5.0 & 12.9 & 18.1 & 478.2 & 0.094 \\
    BootEA~\cite{BootEA18} & 32.3 & 49.9 & 57.9 & 205.5 & 0.410 & 23.4 & 37.4 & 44.5 & 272.1 & 0.307 \\
    SEA~\cite{SEA19} & 17.0 & 33.5 & 42.5 & 191.9 & 0.255 & 14.1 & 28.7 & 37.1 & 207.2 & 0.218 \\
    IMUSE~\cite{IMUSE19} & 17.6 & 34.7 & 43.5 & 182.8 & 0.264 & 8.1 & 19.2 & 25.7 & 397.6 & 0.142 \\
    HyperKA~\cite{HyperKA20} & 13.7 & 24.9 & 30.7 & 712.2 & 0.195 & 16.9 & 29.5 & 34.9 & 738.0 & 0.232 \\
    DualAMN*~\cite{DualAMN21}& 55.0 & {71.4} & {77.1} & {166.6} & {0.627} & \underline{51.6} & \underline{68.3} & \underline{74.4} & 115.1 & \underline{0.594} \\
    \Xhline{1px}
    PoE~\cite{MMKG} & 12.0 & - & 25.6 & - & 0.167 & 10.9 & - & 24.1 & - & 0.154 \\
    MMEA~\cite{MMEA20} & 26.5 & 45.1 & 54.1 & 124.8 & 0.357 & 23.4 & 39.8 & 48.0 & 147.4 & 0.317 \\
    ACK-MMEA$\bullet$ ~\cite{ACKMMEA} & 30.4 & - & 54.9 & - & 0.387 & 28.9 & - & 49.6 & - & 0.360 \\
    MCLEA$\bullet$~\cite{MCLEA} & 44.5 & - & 70.5 & - & 0.534 & 38.8 & - & 64.1 & - & 0.474 \\
    EVA~\cite{EVA20} & 55.6 & 66.6 & 71.6 & 140.0 & 0.609 & 10.3 & 21.7 & 27.8 & 616.8 & 0.164 \\
    MEAformer$\bullet$~\cite{MEAformer} & 57.8 & - & 81.2 & - & 0.661 & 44.4 & - & 69.2 & - & 0.529 \\
    MSNEA~\cite{MSNEA} & \underline{65.3} & \underline{76.8} & \underline{81.2} & \underline{54.0} & \underline{0.708} & 44.3 & 62.6 & 69.8 & \underline{85.1} & 0.529 \\ 
    \Xhline{1px}
    \Method{}$_{\text{GCN-Align}}$ & 55.0 & 71.3 & 77.0 & 66.4 & 0.626 & 47.6 & 64.9 & 71.4 & 80.7 & 0.558 \\ 
    \Method{}$_{\text{(w/o iteration)}}$ & 80.1 & 87.5 & 90.0 & 27.2 & 0.836 & 74.8 & 83.9 & 86.8 & 44.8 & 0.790 \\ 
    \textbf{\Method{}} & \textbf{87.7} & \textbf{93.0} & \textbf{94.5} & \textbf{22.7} & \textbf{0.902} & \textbf{80.5} & \textbf{88.0} & \textbf{90.1} & \textbf{36.4} & \textbf{0.839} \\ 
    Improvements over best baselines& \textbf{22.4}& \textbf{16.2}& \textbf{13.3}& \textbf{31.3}& \textbf{0.194}& \textbf{28.9} & \textbf{19.7} & \textbf{15.7} & \textbf{48.7} & \textbf{0.245} \\ 
    \Xhline{1px}
    \end{tabular}}
    \vspace{-2mm}
\end{table*}
\subsection{Experimental Setup}
\label{sec:exp_setup}
We describe the experimental setup, including datasets, baselines, and evaluation metrics. 

\smallsection{Datasets}
For evaluating the multi-modal ability of \Method{}, we use the following datasets.

\begin{itemize}[topsep=0pt,itemsep=0pt,parsep=0pt,partopsep=0pt,leftmargin=*]
\item To evaluate \Method{} on the \textbf{relational, visual and attribute modalities}, we use
two real-world multi-modal datasets, \FBDB{} and \FBYG{}, provided by~\cite{MMKG, MMEA20}. These datasets are most representative in multi-modal EA task~\cite{MSNEA, MMEA20, MEAformer, ACKMMEA}, and contains the \emph{relational}, \emph{visual}, and \emph{attribute} modalities. For all experiments, we use 20\% of the total alignment pairs as the training set (seed alignment), and use the rest of the pairs as the test set. Table~\ref{tab:datasets} shows the statistics of the multi-modal datasets.

\item 
The above-mentioned datasets do not contain \textbf{temporal modality}. Thus,
for evaluating \Method{} on the ability of modeling \emph{temporal modality}, 
we use two real-world Temporal KG datasets provided by~\cite{TEAGNN21, TREA22}. They are extracted from ICEWS05-15~\cite{ICEWS05-15}, YAGO~\cite{YAGO}, and Wikidata~\cite{Wikidata14}, denoted as \DICEWS{}, \WIKIYAGO{}, respectively. \DICEWS{} has two settings on seed alignment ratios, \DICEWS{-200} and \DICEWS{-1K} that use 200 and 1,000 seed alignment pairs, respectively. Similarly, \WIKIYAGO{} has \WIKIYAGO{-1K}, \WIKIYAGO{-5K}, that use 1,000, and 5,000 seed alignment pairs, respectively. Table~\ref{tab:time_dataset} shows the statistics of the two datasets.
\end{itemize}

\smallsection{Baselines}
We compare \Method{} with the following baselines, which are categorized into three groups: Traditional methods, multi-modal methods and Time-aware methods.

\noindent
\emph{Traditional Methods} are those methods that only use the relational modality (some~\cite{GCN-Align18, IMUSE19} also use the attribute modality). They include:
\begin{itemize}
[topsep=0pt,itemsep=0pt,parsep=0pt,partopsep=0pt,leftmargin=*]
    \item MTransE~\cite{MTransE17}: Embeds different knowledge graphs separately in distinct embedding spaces and provides transitions for entities to map aligned entities.
    \item IPTransE~\cite{IPTransE17}: Iteratively adds newly aligned entity pairs to a set for soft alignment and employs a parameter sharing strategy for different knowledge graphs.
    \item GCN-Align~\cite{GCN-Align18}: Combines structural and attribute information using graph convolutional networks.
    \item BootEA~\cite{BootEA18}: Iteratively labels likely alignments as training data and reduces error accumulation during iterations through an alignment editing method.
    \item SEA~\cite{SEA19}: Utilizes both labeled entities and abundant unlabeled entity information for alignment and incorporates the difference in point degrees with adversarial training.
    \item IMUSE~\cite{IMUSE19}: Employs a bivariate regression model to learn weights for relation and attribute similarities for combining alignment results.
    \item HyperKA~\cite{HyperKA20}: Introduces a hyperbolic relational graph neural network to explore low-dimensional hyperbolic embeddings for entity alignment.
    \end{itemize}

\noindent
\emph{Multi-modal Methods} are those methods that use relational, attribute, and visual modalities. They include: 
\begin{itemize}
[topsep=0pt,itemsep=0pt,parsep=0pt,partopsep=0pt,leftmargin=*]
    \item PoE~\cite{MMKG}: Judges SameAs links between aligned entities by assigning a high probability to true triples and a low probability to false triples. Defines the overall probability distribution as the product of all uni-modal experts.
    \item MMEA~\cite{MMEA20}: Generates entity representations of relational, visual, and numerical knowledge, and then migrates them.
    \item MCLEA~\cite{MCLEA}: Proposes a multi-modal contrastive learning framework to learn the representations of entities from different modalities.
    \item MSNEA~\cite{MSNEA}: Introduces a multi-modal siamese network approach to learn the representations of entities from different modalities.
    \item MEAformer~\cite{MEAformer}: Proposes a multi-modal transformer-based approach to learn the representations of entities from different modalities.
    \item ACK-MMEA~\cite{ACKMMEA}: Proposes an attribute-consistent knowledge graph entity alignment approach considering multiple modalities.
\end{itemize}

\noindent
\emph{Temporal modality Methods} are those methods that specifically consider both relational and temporal modalities. They are

\begin{itemize}
[topsep=0pt,itemsep=0pt,parsep=0pt,partopsep=0pt,leftmargin=*]
    \item TEA-GNN~\cite{TEAGNN21}: As the first time-aware EA model, it models temporal information as embeddings.
    \item TREA~\cite{TREA22}: Uses a temporal relation attention mechanism to model temporal information.
    \item STEA~\cite{cai2022simple}: Constructs a time similarity matrix by a dictionary counting the relation between timestamps and entities.
    \item DualMatch~\cite{DualMatch}: Proposes to separately align entities using relational and temporal information, and then fuse the results.
    \end{itemize}


\begin{table}[t]\small
\centering
\caption{The results on \DICEWS{} and \WIKIYAGO{}.}
\label{table:time}
\vspace{-2mm}
    \setlength{\tabcolsep}{1.2mm}
{
\begin{tabular}{l|ccc|ccc}
\Xhline{1px}
\multirow{2}{*}{Method} & \multicolumn{3}{c|}{\DICEWS{-200}} & \multicolumn{3}{c}{\DICEWS{-1K}}\\
& H@1$\uparrow$ & H@10$\uparrow$ & MRR$\uparrow$ & H@1$\uparrow$ & H@10$\uparrow$ & MRR$\uparrow$\\
\Xhline{1px}
TEA-GNN~\cite{TEAGNN21} & 87.6 &  94.1 &  0.902 & 88.7 & 94.7 & 0.911\\
TREA~\cite{TREA22} &  91.0 &  96.0 & 0.927 & 91.4 & 96.6 & 0.933\\
STEA~\cite{cai2022simple} & 94.3 &  96.8 & 0.954 & 94.5 & 96.7 &  0.954\\
DualMatch~\cite{DualMatch} & \textbf{95.3} &  \textbf{97.4} &  \textbf{0.961} & \textbf{95.3} & \textbf{97.3} & \textbf{0.961}\\
\Method{}& \textbf{95.3} & 97.2 & \textbf{0.961} & 95.0 & 97.0 & 0.958\\
\Xhline{1px}

\multirow{2}{*}{Method} &\multicolumn{3}{c|}{\WIKIYAGO{-1K}}& \multicolumn{3}{c}{\WIKIYAGO{-5K}}\\
& H@1$\uparrow$ & H@10$\uparrow$ & MRR$\uparrow$ & H@1$\uparrow$ & H@10$\uparrow$ & MRR$\uparrow$ \\
\Xhline{1px}
TEA-GNN~\cite{TEAGNN21} &  72.3 & 87.1 & 0.775 & 87.9 & 96.1 &  0.909\\
TREA~\cite{TREA22} & 84.0 &  93.7 & 0.885 & 94.0 & 98.9 &  0.958\\
STEA~\cite{cai2022simple}&94.3 & \textbf{98.9} & 0.962 & 96.1 &  99.2 & 0.974 \\
DualMatch~\cite{DualMatch} &  94.7 & 98.4 &  0.961 & 
\textbf{98.1} & \textbf{99.6} & \textbf{0.986}\\
\Method{}& \textbf{96.4} & 98.4 & \textbf{0.972} & 97.7 & \textbf{99.6} & 0.984\\
\Xhline{1px}
\end{tabular}
}
\end{table}

\smallsection{Evaluation Metrics}
We use the widely-adopted Hits@N (H@N, in percentage), Mean Reciprocal Rank (MRR) and Mean Rank (MR) metrics to evaluate the performance of \Method{} and the baselines. 
\HitNFull{} (in percentage) denotes the proportion of correctly aligned entities in the top-$N$ ranks.
MRR is the average of the reciprocal ranks of the correctly aligned entities, where reciprocal rank reports the mean rank of the correct alignment.
MR is the average of the ranks of the correctly aligned entities.
Higher \HitN{}, MRR, and lower MR indicates higher EA accuracy.

\smallsection{Implementation Details}
We choose several state-of-the-art models as the backbone models for each modality.
We choose DualAMN~\cite{DualAMN21} to learn the knowledge graph relations, SwinTransformer~\cite{SwinTransformer} to encode images into vectors, RoBERTa~\cite{RoBERTa19} to encode attribute names into vectors. 
We iteratively refine the relational model for 3 iterations, with the hyper-parameters the same as~\cite{DualAMN21}. We allow one entity to have a maximum of 6 images. We set $k=10$ for the Sinkhorn algorithm. We implement all our codes in Python and run all experiments on a server with 8 NVIDIA RTX 3090 GPUs.

\subsection{Main Results}
\smallsection{Modalities and Evaluation Results}
We evaluate \Method{} on the \FBDB{} and \FBYG{} datasets, considering relational, visual, and attribute modalities. Results are presented in Table~\ref{tab:main_table} with * denoting our reproduced baselines, $\bullet$ denoting results from original papers, and other results from~\cite{MSNEA}. The best results are \textbf{bold}, and the best among baselines are \underline{underlined}. Our observations are as follows:

First, \Method{} outperforms the baselines significantly. On \FBDB{}, \Method{} surpasses the best baseline by \textbf{22.4\%} in Hits@1 and \textbf{0.194} in MRR. On \FBYG{}, \Method{} outperforms the best baseline by \textbf{28.9\%} in Hits@1 and \textbf{0.245} in MRR, demonstrating its effectiveness for multi-modal EA. 
Second, interestingly, some recent multi-modal approaches perform worse than the single-modal baseline DualAMN~\cite{DualAMN21}. These approaches fail to effectively fuse information from different modalities. In contrast, \Method{} successfully fuses information and outperforms the baselines by a large margin.

\smallsection{Temporal and Relational Modalities}
We evaluate \Method{} on the \DICEWS{} and \WIKIYAGO{} datasets, focusing on temporal and relational modalities. Results in Table~\ref{table:time} show that \Method{} outperforms two temporal-aware EA models~\cite{TEAGNN21, TREA22} by a significant margin, despite not explicitly modeling the temporal modality. Additionally, \Method{} is comparable to the state-of-the-art temporal-aware EA model DualMatch~\cite{DualMatch}, while being a general framework applicable to different modalities. DualMatch, on the other hand, is specifically designed for temporal-aware EA and cannot be applied to other modalities.

\begin{figure*}[t]
    \centering
    \includegraphics[width=6.7in]{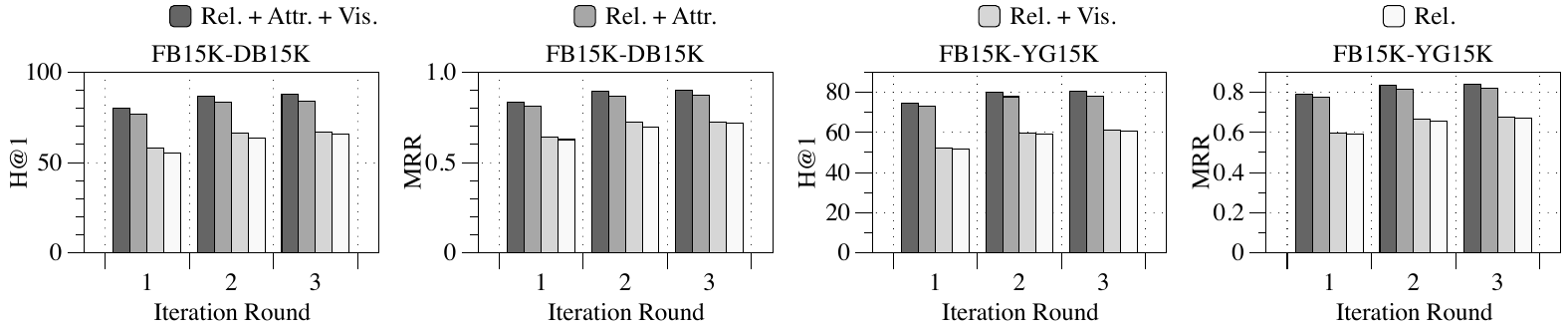}
    \caption{H@1 and MRR vs. Iteration Round}
    \vspace*{-2mm}
    \label{fig:iteration}
\end{figure*}

\subsection{Ablation Study}

We present various variants of \Method{} in Table~\ref{tab:main_table}. \Method{}$_{\;\text{GCN-Align}}$ refers to \Method{} with GCN-Align~\cite{GCN-Align18}, a classic EA model, as its relational model. \Method{}$_{\;\text{(w/o iteration)}}$ refers to \Method{} without iterative refinement of the relational model.
First, each variant exhibits performance degradation, demonstrating the effectiveness of each component of \Method{}.
Second, even without iterative refinement, our proposed \Method{} outperforms all previous methods.
Third, while \Method{}$_{\;\text{GCN-Align}}$ outperforms GCN-Align~\cite{GCN-Align18}, it is still inferior to some baselines. This highlights the importance of the relational modality in multi-modal EA.

We also offer a comprehensive analysis of the impact of removing individual modalities, including Attribute, Visual, and all side modalities in each iteration, from \Method{}, as depicted in Figure~\ref{fig:iteration}. It is clear that removing any of these modalities results in a decline in performance, highlighting the effectiveness of each modality. Additionally, the absence of the iterative refinement process leads to a decrease in accuracy, further demonstrating the effectiveness of \Fuse{}.

\subsection{Study of \Path{} on Different Modalities}
As mentioned in Section~\ref{sec:method}, \Path{} is a general framework that can be applied to different modalities to obtain the alignment matrices. To evaluate the effectiveness of \Path{} on different modalities, we assess the alignment matrices generated by \Path{} for each modality. First, we apply \Path{} to each modality to obtain the alignment matrices and then use these matrices to perform zero-shot EA.
We evaluate \Method{} for its zero-shot EA ability on different modalities by directly evaluating the alignment matrices obtained by \Path{} on each modality. In each setting, \Path{} is only allowed to use the corresponding modality.

\smallsection{Attribute modality}
We evaluate \Path{} on its EA ability for the attribute modality by establishing a variant with only attribute modality information (referred to as \Path{} (Attr.)). We compare \Path{} (Attr.) with ACK-MMEA~\cite{ACKMMEA}, a baseline introduced in 2023 that leverages all accessible modalities.
We also establish a baseline that encodes both attribute names and values as a whole sentence using language models and then calculates the similarity of these "sentence" embeddings (denoted as AttrSim). The results are presented in Table~\ref{table:attr}. It can be observed that \Path{} outperforms AttrSim by a significant margin. Furthermore, \Path{} (Attr.) outperforms ACK-MMEA even using \textbf{only} the attribute modality, with a significant 15\% relatively improvement on MRR of \FBYG{} dataset. This demonstrates the effectiveness of \Path{} on the attribute modality. Lastly, AttrSim performs better on the \FBYG{} dataset because YAGO has limited attribute types, making it easier to distinguish the attribute values.

\begin{figure}[t]
    \centering
    \includegraphics[width=3.4in]{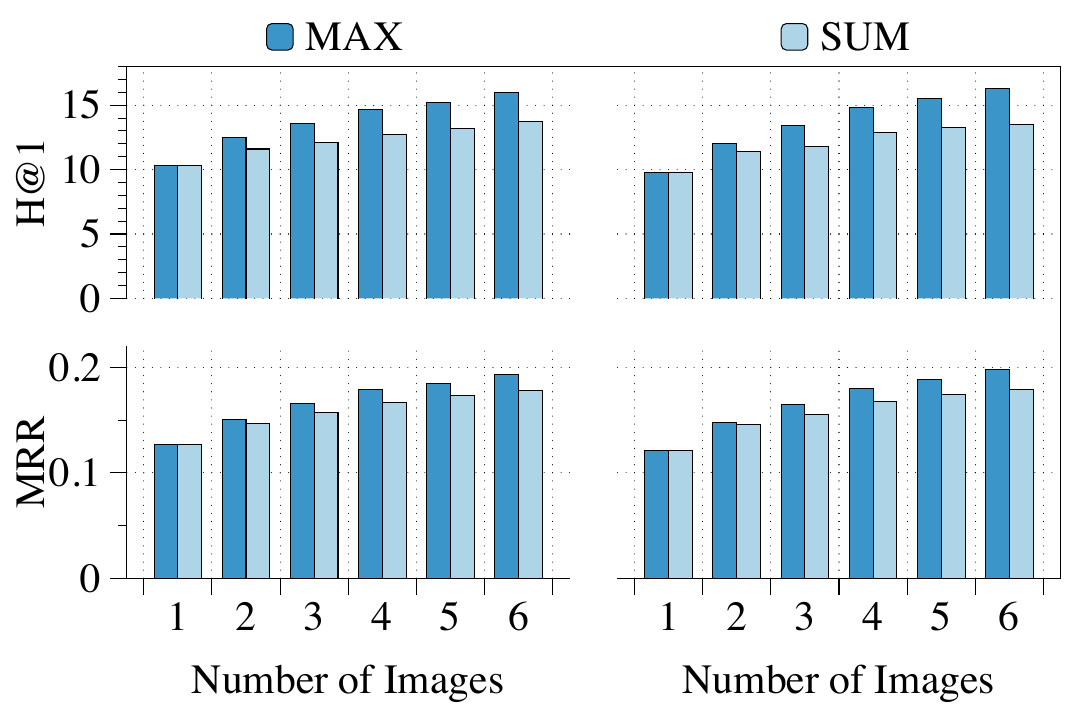}
    \vspace*{-22mm}
    \\
    \hspace{0.2in}
    \subfigure[FB15K-DB15K]{
    \includegraphics[width=1.4in]{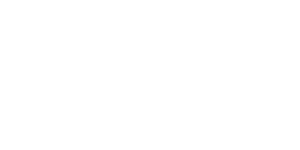}
    }\hspace{-5mm}
    \subfigure[FB15K-YG15K]{
    \includegraphics[width=1.4in]{figs/transparent.png}
    }
    \vspace*{-1mm}
    \caption{H@1 and MRR vs. Number of Images}
    \vspace*{-4mm}
    \label{fig:image_cnt}
\end{figure}

\begin{table}[t]
\centering
\caption{The results of Attribute Modality. ACK-MMEA utilizes all the accessible modalities, while AttrSim and \Path{} (Attr.) only utilize the attribute modality.}
\label{table:attr}
\vspace{-3mm}
    \setlength{\tabcolsep}{1.2mm}
{
\begin{tabular}{l|ccc|ccc}
\Xhline{1px}
\multirow{2}{*}{Method} & \multicolumn{3}{c|}{\FBDB{}} & \multicolumn{3}{c}{\FBYG{}}\\
& H@1$\uparrow$ & H@10$\uparrow$ & MRR$\uparrow$ & H@1$\uparrow$ & H@10$\uparrow$ & MRR$\uparrow$\\
\Xhline{1px}
ACK-MMEA & 30.4 &  \textbf{54.9} & \textbf{0.387} & 28.9 & 49.6 & 0.360 \\
AttrSim &3.7&11.4&0.064&10.7&23.9&0.153 \\
\Path{} (Attr.) &\textbf{32.4}&51.2&0.384&\textbf{34.7}&\textbf{57.0}&\textbf{0.416}\\
\Xhline{1px}
\end{tabular}
}
\end{table}

\smallsection{Visual modality}  In our dataset, it's possible for a single entity to be associated with multiple images. For the FB15K dataset, the images are sourced from the image links provided by MMKB~\cite{MMKG}. In the cases of YG15K and DB15K, we gather images from Wikimedia. In our experiments, we've chosen to utilize the first six images for each entity. However, if there are more than six images associated with a particular entity, we cap the selection at six.


Here, we vary the number of images from 1 to 6 in \Path{} and evaluate the performance of \Method{} on the \FBDB{} and \FBYG{} datasets using only the visual modality. We also compare the results obtained using both the max and sum operators of \Path{}. The results are presented in Fig.~\ref{fig:image_cnt}. As expected, with more images, the performance of \Path{} gradually improves. Additionally, using the max operator consistently outperforms the sum operator. This is because the max operator better captures the most important information from the images.
\section{Conclusions}

This paper presents a novel method called \Method{} for multi-modal EA in KGs, mitigating the challenges of inconsistent and inefficient modality modeling.
\Method{} addresses the limitations of incompleteness and overlapping entities in KGs by incorporating additional modalities, enhancing the effectiveness and robustness of 
\Method{} first proposes \Path{}, introducing a unified modeling technique that simplifies the alignment process by constructing paths connecting entities and modality nodes to model multiple modalities. This enables direct transformation of information into alignment results. Furthermore, \Method{} introduces \Fuse{}, an effective fusion method that iteratively fuses information from different modalities using the path as an information carrier.
The experimental results demonstrate the superiority of \Method{} over state-of-the-art methods in multi-modal EA. Specifically, \Method{} achieves more accurate and comprehensive entity alignment, with $22.4\%-28.9\%$ absolute improvement on Hits@1, and $0.194-0.245$ absolute improvement on MRR. 
The use of \Path{} and \Fuse{} contributes to a more holistic representation of entities across different KGs, leading to improved alignment results.

\section*{Limitations}

\smallsection{Scalability} \Method{} has a quadratic complexity due to the nature of EA. This is a common problem for EA methods, as we need to compare each entity pair from different KGs. 
Also, there lack of a large-scale multi-modal EA dataset for evaluation. 
Some recent works~\cite{LargeEA22, ClusterEA, LightEA} have proposed to use of approximate methods to reduce the number of entity pairs. However, these methods do not apply to \Method{} because they are designed mostly for relational EA, and inevitably hurt the EA accuracy as a tradeoff. Also, there are large-scale GNN training and inference infrastructures that can be applied to EA~\cite{wang2022neutronstar, yin2022dgi, tan2023quiver, wang2023hytgraph} We leave the scalability issue for future work.

\smallsection{Modality Selection} \Method{} requires the modality nodes to be explicitly provided. However, in real-world applications, the modality nodes are not always available. We leave the modality selection problem for future work. Also,  \Method{} does not consider textual modality, which is an important modality in KGs. This is due to the potential test data leakage problem. We leave resolving the problem for future work.


\smallsection{Aligning KGs with different modality distributions} \Method{} requires the two KGs to align have roughly similar modality distributions. This is because \Method{} uses the modality nodes to guide the alignment process. However, in real-world applications, the two KGs may have different modality distributions. We leave the problem for future work.


\balance
\bibliographystyle{ACM-Reference-Format}
\bibliography{REFER}
\pagebreak

\end{document}